\documentclass[lettersize,journal]{IEEEtran}

\usepackage{amsmath,amsfonts,amssymb}
\usepackage{algorithmic}
\usepackage{algorithm}
\usepackage{array}
\usepackage[caption=false,font=normalsize,labelfont=sf,textfont=sf]{subfig}
\usepackage{textcomp}
\usepackage{stfloats}
\usepackage{url}
\usepackage{verbatim}
\usepackage{graphicx}
\usepackage{cite}
\usepackage{comment}
\usepackage[colorlinks]{hyperref}

\usepackage{booktabs}
\usepackage{enumerate}
\usepackage{enumitem}
\usepackage{makecell}
\usepackage{tabu}
\usepackage{multirow}
\usepackage{utfsym}
\usepackage{colortbl} 

\newcommand{\ie}{\textit{i}.\textit{e}.\ }
\newcommand{\eg}{\textit{e}. \textit{g}.\ }

\begin{document}

\title{CM-MaskSD: Cross-Modality Masked Self-Distillation for Referring Image Segmentation}

\author{Wenxuan Wang, Xingjian He, Yisi Zhang, Longteng Guo, Jiachen Shen, 
Jiangyun Li\textsuperscript{\dag}, Jing Liu, \IEEEmembership{Member, IEEE} 
\thanks{This work was supported by the National Key Research and Development Program of China (No.2022ZD0118801), National Natural Science Foundation of China (U21B2043,62102416), in part by the Natural Science Foundation of China under Grant 42201386, the International Exchange Growth Program for Young Teachers of USTB under Grant QNXM20220033, Scientific and Technological Innovation Foundation of Shunde Graduate School, USTB (BK20BE014).}
\thanks{W. Wang, X. He, L. Guo and J. Liu are with the Institute of Automation, Chinese Academy of Sciences, Beijing 100190, China, while W. Wang and J. Liu are also with the School of Artificial Intelligence, University of Chinese Academy of Science, Beijing 100190, China. (e-mail: wangwenxuan2023@ia.ac.cn, xingjian.he@nlpr.ia.ac.cn, jliu@nlpr.ia.ac.cn, longteng.guo@nlpr.ia.ac.cn)}
\thanks{Y. Zhang, J. Shen and J. Li are with the University of Science and Technology Beijing, Beijing 100083, China. (e-mail: leejy@ustb.edu.cn)}
\thanks{\textsuperscript{\dag}Corresponding author.}
}

\maketitle

\begin{abstract}

Referring image segmentation (RIS) is a fundamental vision-language task that intends to segment a desired object from an image based on a given natural language expression. Due to the essentially distinct data properties between image and text, most of existing methods either introduce complex designs towards fine-grained vision-language alignment or lack required dense alignment, resulting in scalability issues or mis-segmentation problems such as over- or under-segmentation. To achieve effective and efficient fine-grained feature alignment in the RIS task, we explore the potential of masked multimodal modeling coupled with self-distillation and propose a novel cross-modality masked self-distillation framework named CM-MaskSD, in which our method inherits the transferred knowledge of image-text semantic alignment from CLIP model to realize fine-grained patch-word feature alignment for better segmentation accuracy. Moreover, our CM-MaskSD framework can considerably boost model performance in a nearly parameter-free manner, since it shares weights between the main segmentation branch and the introduced masked self-distillation branches, and solely introduces negligible parameters for coordinating the multimodal features. Comprehensive experiments on three benchmark datasets (\ie RefCOCO, RefCOCO+, G-Ref) for the RIS task convincingly demonstrate the superiority of our proposed framework over previous state-of-the-art methods.

\end{abstract}


\section{Introduction}
\label{introduction}

\IEEEPARstart{R}{eferring} image segmentation (RIS) aims to segment specific regions of input images corresponding to the given language expression. 
RIS has become one of the most challenging vision-language tasks due to its requirement for mutual understanding across the two different modalities. 
In addition, compared with conventional single-modality (\ie image or video) segmentation, RIS partly resolves the limitations of segmentation targets solely on predefined categories. 
Considering that diverse targets are required in real-world downstream tasks, RIS could potentially be employed in a wide range of applications, including language-based human-object interaction and interactive image editing.

\begin{figure}[t]
    \centering \includegraphics[width=0.48\textwidth]{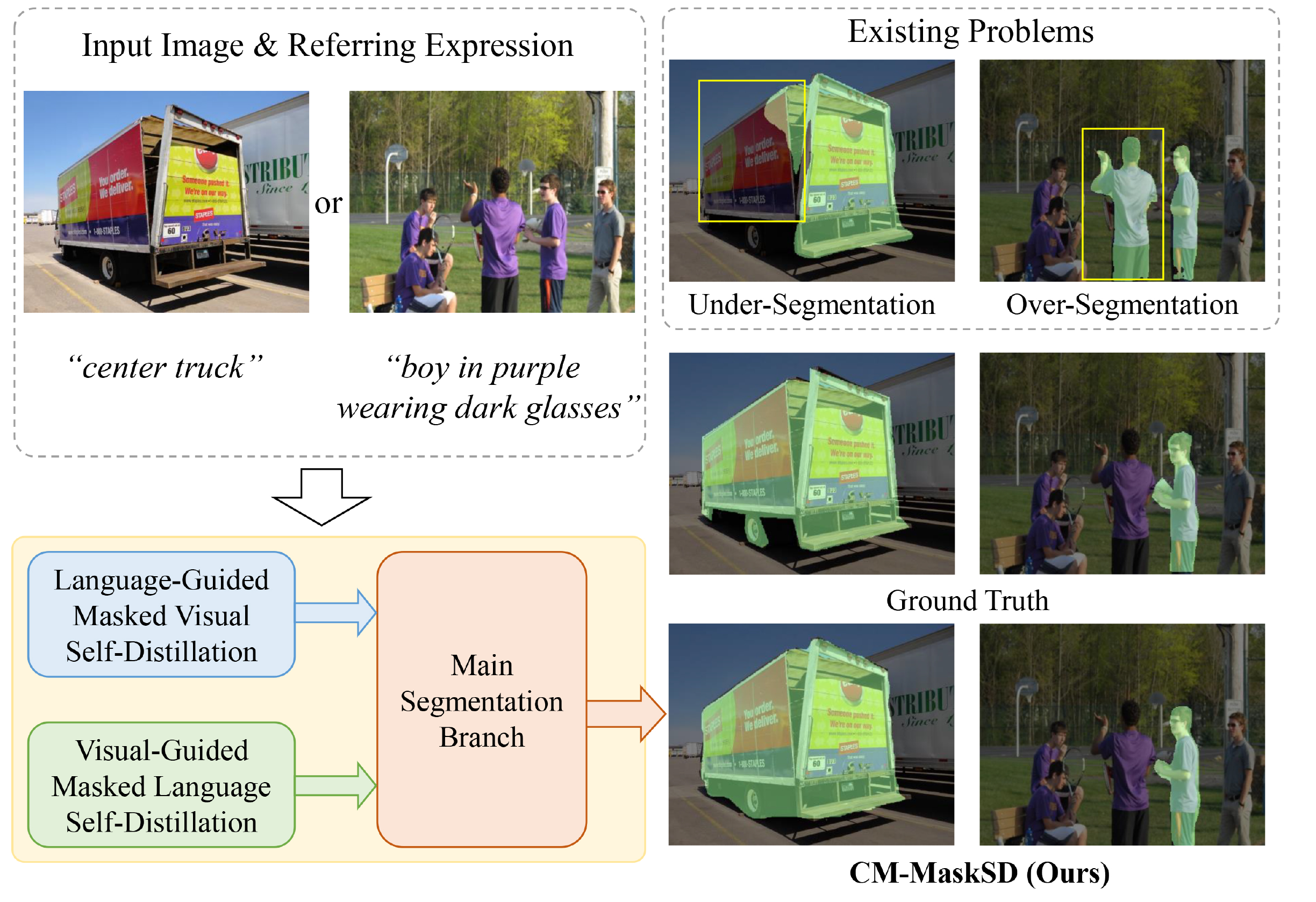}
    \vspace{-10pt}
    \caption{
    The illustration of our pipeline for referring image segmentation task.
    }
    \vspace{-15pt}
    \label{figure_intro}
\end{figure}

Since the concept of RIS task was initially proposed in \cite{hu2016segmentation}, different multimodal frameworks for referring image segmentation have been designed to deal with the feature extraction and interaction between visual and linguistic modalities, including the first attempt of introducing recurrent LSTM network \cite{hu2016segmentation}, CLIP-driven framework \cite{wang2022cris}, language-aware vision Transformer network \cite{yang2022lavt}, and convolution-free network \cite{kim2022restr}.  
However, due to the considerable differences between visual and language modalities, feature alignment has become a crucial challenge for precise segmentation. 
Previous works have made great efforts to overcome this challenge.
For instance, CRIS \cite{wang2022cris} achieves semantic consistency by propagating semantic information from textual representations to each image pixel, yet it neglects to correlate the important visual information with language representations. 
LAVT \cite{yang2022lavt} enhances the model's capability of cross-modality alignment through a multi-stage structure, yet such a sophisticated network is inflexible for further improving fine-grained feature alignment. 
It is notable that without aligning cross-modality information correctly, models are prone to the mis-segmentation problem (\ie over- and under-segmentation).
As shown in Fig. \ref{figure_intro}, models could not generate a complete segmentation mask of the main parts without adequate exploration of the correlation between image regions and their relevant words, leading to the under-segmentation problem. 
Besides, image regions with essentially weak relevance to the whole expression may be wrongly guided by some misleading words in the given text, resulting in the over-segmentation problem.
As shown in the second column on the right side of Fig. \ref{figure_intro}, rather than generating the segmentation mask of the specific boy wearing dark glasses, CRIS \cite{wang2022cris} has segmented redundant parts (\ie another boy in purple without wearing glasses) since the less important words (boy in purple) are wrongly considered as the most relevant with the image.
Although advanced models could be adopted for comprehensive feature extraction of the given images and texts, a novel generic framework is expected to overcome the aforementioned two drawbacks and effectively realize more fine-grained feature alignment.

Therefore, to better assist models in mastering the capability of fine-grained feature alignment, we explore the potential of masked self-distillation applying to referring image segmentation task for the first time. 
In this paper, we provide a \textbf{C}ross-\textbf{M}odality \textbf{Mask}ed \textbf{S}elf-\textbf{D}istillation framework (CM-MaskSD) to densely correlate the multimodal features in a simple and efficient manner. 
As shown in Fig. \ref{figure_intro}, our proposed framework enables effective mutual guidance between visual and textual modalities by taking advantage of the bidirectional cross-modality interaction. 
Except for the main branch for realizing the segmentation task, two symmetric branches are additionally designed for masked self-distillation. 
{In essence, we intend to introduce a bidirectional cross-modality guided masking strategy to mask the feature vectors that may lead to incorrect segmentation results due to erroneous cross-modal feature correlations.
By constraining the segmentation results before and after masking to be consistent, the knowledge distillation implicitly occurs between the main segmentation branch and the introduced two masked self-distillation branches—the model implicitly reinforces the cross-modality features that should be strongly correlated, while attenuating those that should be weakly associated. This facilitates a finer-grained alignment of cross-modality visual and textual features, aiming to minimize the occurrence of segmentation errors (i.e. over-segmentation and under-segmentation). 
To further pursue the highly efficient architecture, the weights of both the main segmentation branch and the two distillation branches are shared together to greatly reduce the introduced parameters.
In this way, since both the teacher and student model in the traditional distillation pipeline here correspond to the same model weights in our architecture, which essentially does not include the conventional knowledge is transferred from one structure to another, the term ``self-distillation" is derived from this structural design.}

Specifically, the first branch is the language-guided masked visual self-distillation branch, in which we introduce the textual global features from the language encoder to guide the masked visual self-distillation. 
The correlation vector is firstly calculated between the visual feature embeddings of image patches and the textual global token, then the TopK text-related image patches that are most related to textual global representations are selected. 
{After this correlation filtering operation, the relatively irrelevant visual tokens with low correlation values are filtered out and the strongly correlated visual tokens are saved for the following masked self-distillation design.
In this manner, the subsequent masking strategy will be more effective and targeted in addressing erroneous cross-modal feature associations that can lead to mis-segmentation.}
Then we randomly mask the embedded features of TopK related image patches with a masking ratio $\alpha$ and send all the resulted visual tokens into the following model for segmentation.
To implicitly guide model to achieve dense alignment between visual and textual features, the optimization target is conducted by pulling closer the segmentation results of masked visual self-distillation branch and the main segmentation branch (\ie without any masking).  
Similarly, the visual-guided masked language self-distillation branch is symmetric to the aforementioned masked visual self-distillation branch. 
Finally, by jointly employing the bilateral masked self-distillation branches, our framework could realize more fine-grained multimodal feature alignment, hence accomplishing the referring image segmentation task in a more precise manner. 
Noticeably, based on the main segmentation branch (\ie a strong baseline), our masked self-distillation design solely introduces negligible parameters yet can greatly boost model performance via sharing weights between the self-distillation branches and the main branch.
The experimental results clearly show that our framework achieves superior performance over previous state-of-the-art (SOTA) methods on the three benchmark datasets for RIS task.

Our main contributions can be summarized as follows:
\setlist{nolistsep}
\begin{itemize}[noitemsep,leftmargin=*]
    \item 
    We present the first study to explore the powerful potential of masked multimodal modeling with self-distillation for RIS task. Our proposed novel framework CM-MaskSD can inherit the transferred knowledge of image-text semantic alignment from CLIP model to realize the dense text-patch feature alignment for higher segmentation accuracy. 
    \item By fully taking advantage of the proposed correlation filtering mechanism and cross-modality guided masking strategy in our dual masked self-distillation branches, our method can effectively achieve more fine-grained vision-language feature alignment, which is crucial for the RIS task. 
    \item Our CM-MaskSD is a simple yet effective and generic framework,  
    among which our masked self-distillation design is essentially plug-and-play and easy-to-implement.
    \item The experimental results on the three benchmark datasets for referring image segmentation convincingly demonstrate the superiority of our CM-MaskSD over previous state-of-the-art methods. Moreover, via sharing weights, our framework can consistently boost model performance with only introduced negligible parameters and none extra computational costs for inference.
\end{itemize}

\section{Related Work}
\label{relatedwork}
\noindent \textbf{Referring Image Segmentation} is to generate the category masks of target objects in an image according to the given natural language description. Since the input consists of multimodal information, constructing an effective framework for feature modeling and interaction between textual and visual features is considered the most crucial part of the entire task. 
The RIS task is first brought up in  \cite{hu2016segmentation}, which simply concatenates linguistic and visual features extracted by LSTM \cite{hochreiter1997long} and convolutional neural network separately, and predicts the final segmentation mask through a fully connected network. 
Some of the subsequent works \cite{shi2018key, margffoy2018dynamic, li2018referring, qiu2019referring, ye2020dual, shi2020query, lin2021structured} follow the paradigm of modeling text expression and image features independently, and then set fusion pipeline to introduce language information into pixel-level activation. 
However, \cite{liu2017recurrent} believes that joint modeling is more intrinsic for RIS task and proposes the multimodal ConvLSTM to encode visual information, spatial cues and the sequential interaction between each word. 
Nevertheless, rather than regarding the input sentence as an individual unit, MattNet \cite{yu2018mattnet} treats it as a hybrid of objects' position, appearance and relationship to others, putting forward a two-stage network to select generated regions of interest with textual guidance.
Different from the aforementioned methods that employ implicit feature interaction and fusion between visual and linguistic modalities, \cite{huang2020referring} tends to adopt a progressive manner, using entity and attribute words to perceive all the entities involved in the expression and further inferring the relationship between entities to highlight the relevant objects of referring expression. 
Besides, motivated by previous application of contrastive learning in language-image pre-training, CRIS \cite{wang2022cris} propagates fine-grained semantic information from text to visual embeddings via a joint visual-language decoder and enhances cross-modality consistency with contrastive learning.

As Transformer and attention mechanisms have been well studied in various linguistic and visual downstream tasks, their advantage over convolutional neural networks in capturing long-range dependency seems apparent.
In CMSA \cite{ye2019cross}, cross-modality self-attention module is employed to obtain long-range dependency between two modalities. 
VLT \cite{ding2021vision} redefines RIS as a direct attention problem, introducing Transformer and multi-head attention to query a given image with the language expression.
ReSTR \cite{kim2022restr} is proposed as the first convolution-free, Transformer-based referring image segmentation model, which only uses different Transformer encoders to realize the respective feature extraction of text and image and the following multiple interactions between them. 
Furthermore, LAVT \cite{yang2022lavt} operates early multimodal feature fusion on the constructed multi-level Transformer encoder, achieving significantly better cross-modality alignment. 
In addition, \cite{wu2022towards} considers negative sentences as inputs to enhance the robustness of model to the misdescription or misleading by the given text. 

\noindent \textbf{Mask Image/Language Modeling} is an effective self-supervised pre-training pattern for learning general representations, which has been studied in the field of natural language processing (NLP) and computer vision (CV). 
BERT \cite{devlin2018bert} and its variants \cite{liu2019roberta} \cite{clark2020electra} yield SOTA performance in a broad range of NLP tasks by introducing masked language modeling (MLM). 
With the success of MLM in the NLP field and the emergence of vision Transformers \cite{dosovitskiy2020image}, BEiT \cite{bao2021beit} and BEiTV2 \cite{peng2022beit} introduce a classifier to predict masked image tokens, which is supervised by the encoded visual patches from offline tokenizer.
SimMIM \cite{xie2022simmim} directly adopts the low-level image features (\ie pixel's RGB value) as prediction targets, leading to considerable performance gains compared to conventional self-supervised pretext tasks. 
Instead of feeding masked tokens as input to the encoder, MAE \cite{he2022masked} develops a straightforward decoder to reconstruct image patches, resulting in a significant decrease in pre-training computational costs.
{To break the limitation that MAE-based methods can only be performed on the standard vision Transformers \cite{dosovitskiy2020image} and explore the potential of masking operation for various downstream tasks, a lot of works \cite{huang2022green,liu2022mixmim,tian2023designing,wang2023positive} have been proposed.}

{Following the great performance obtained by knowledge distillation in many works \cite{chen2021feature,deng2021extended,chen2021temporal,hao2022cdfkd,hu2022mmnet}, a powerful model compression technique, knowledge distillation has also shown impressive performance in masked image modeling}.
Instead of directly imitating the output of teacher network, MGD \cite{yang2022masked} allows the student model to recover teacher model's feature representations with randomly masked feature maps, achieving excellent improvements on various visual downstream tasks. 
Similarly, DMAE \cite{bai2023masked} aligns the intermediate features between teacher model and student model, studying the potential of distilling knowledge from MAE. 
However, considering that not all pixels of feature maps contribute equally to model performance, MasKD \cite{huang2022masked} utilizes masked feature distillation to prompt student model to adaptively learn the values of the teacher model's feature maps at each position based on their informative contribution. 
Such an attention-aware idea can also be directly applied to the masking process. For instance, MaskedKD \cite{son2023maskedkd} leverages the attention maps learned by student model to mask the input image of teacher network, providing a simple yet efficient strategy to reduce the distillation cost of ViT.

Different from the above works, in this paper, we conduct the pioneering exploration on building an effective and also efficient masked self-distillation architecture on both text and image modalities for better accomplishing RIS task. 
Specifically, two symmetric distillation branches are designed to enhance the model's comprehension of mutual correlations between the feature representations of language expression and image. 
Among each branch, cross-modality guided correlation filtering and the following masking operation are performed to promote the expected fine-grained feature alignment.

\section{Methodology}
\label{methodology}

\begin{figure*}[htbp]
    \centering
    \includegraphics[width=0.96\textwidth]{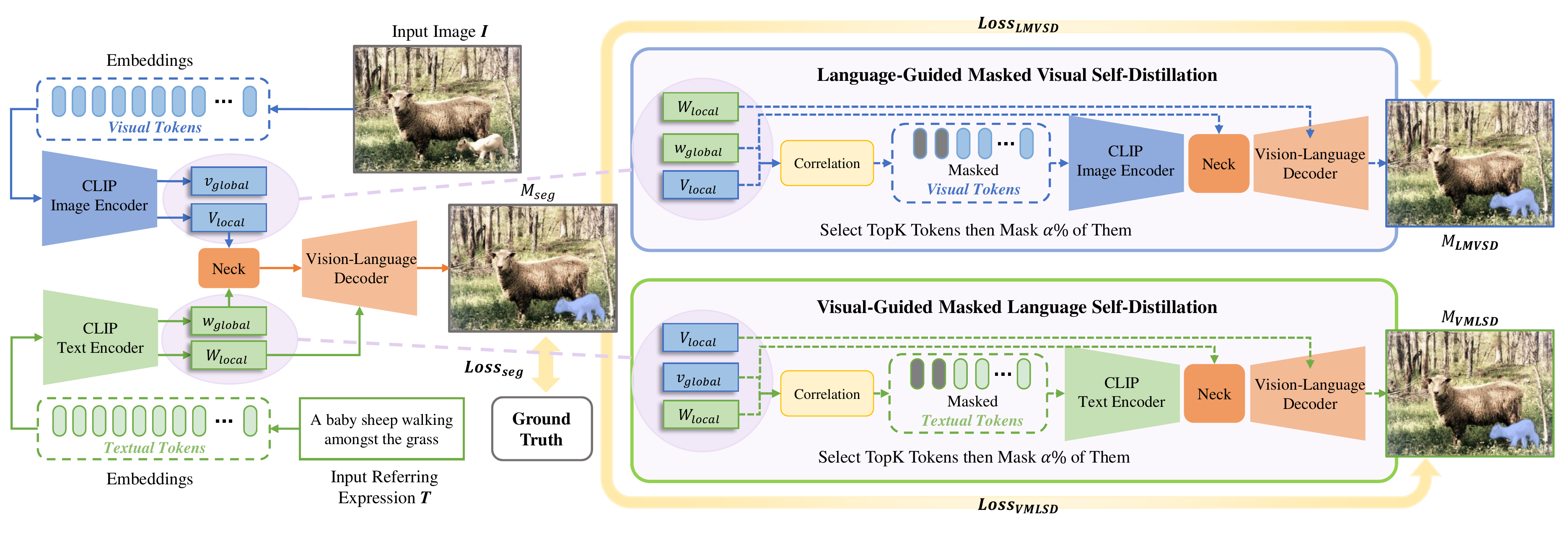}
    \vspace{-10pt}
    \caption{
    The architecture of our CM-MaskSD framework. It consists of a multimodal segmentation branch and two symmetric masked self-distillation branches that are designed for more fine-grained visual and textual feature alignment.
    During training, the main segmentation loss $Loss_{seg}$ coupled with two self-distillation loss $Loss_{LMVSD}$ and $Loss_{VMLSD}$ are jointly employed to pull close the segmentation masks generated by main branch and cross-modality guided masked self-distillation branches.
    For inference, only the main segmentation branch is preserved to acquire the final segmentation masks.
    }
    \label{fig_method}
    \vspace{-10pt}
\end{figure*}

As shown in Fig.~\ref{fig_method}, our Cross-Modality Masked Self-Distillation (CM-MaskSD) framework includes a multi-modal main segmentation branch and two symmetric masked self-distillation branches that are  designed for achieving more fine-grained feature alignment between referring expression and visual representations. 
The details of our CM-MaskSD are presented in the following.

\vspace{-10pt}
\subsection{Main Segmentation Branch}
\label{sec:mainsegmentationbranch}
Since CLIP \cite{radford2021learning} is capable of directly learning transferable visual concepts from large-scale collections of image-text pairs, we leverage the pre-trained weights of CLIP as our segmentation backbone.
To obtain the extracted visual and linguistic feature representations, the image encoder and the text encoder are employed respectively.
Given the input image $I$ and referring text $T$, we first send them to the feature embedding layer to obtain the visual embedding $E_{I}$ and textual embedding $E_{T}$. These embedded feature tokens are then fed to their respective encoders to obtain the visual and textual feature representations, which can be expressed as follows:
\begin{equation}
    \begin{split}
    V_{local}, v_{global} &= \text{CLIP Image Encoder}(E_{I})\\
    W_{local}, w_{global} &= \text{CLIP Text Encoder}(E_{T})
    \end{split}
    \label{eq:clip_encoder}
\end{equation}
where $V_{local}$ and $W_{local}$ are the output sequences of visual and textual tokens respectively, each of which corresponds to a single visual image patch or a textual word. $v_{global}$ denotes the [\texttt{class}] token that serves as the image-level representation with strong semantics. $w_{global}$ is the text-level representation which is the aggregated feature representation of all textual tokens $W_{local}$.
Subsequently, an effective neck module is used to fuse the multimodal feature $F$ by taking $V_{local}$ and $w_{global}$ as input, followed by a vision-language decoder that generates the final segmentation results $M_{seg}$. 
{Specifically, the neck module composes of several linear layers followed by non-linear functions which is utilized to perform cross-modality fusion between hierarchical visual representation and textual representation with an output of multi-modal representation.
And the vision-language decoder follows the standard architecture of Transformer design, in which the multi-modal feature representations sequentially pass through the multi-head self-attention and multi-head cross-attention structure with textual features as key and value.}
Note that the employed neck and vision-language decoder are solely introduced to provide a strong segmentation baseline, which follows the same standard architecture as CRIS~\cite{wang2022cris}.
\begin{equation}
    \begin{split}
    F &= \text{Neck}(V_{local}, w_{global}) \\
    M_{seg} &= \text{Vision-Language Decoder}(F, W_{local})
    \end{split}
    \label{eq:neck_decoder}
\end{equation}

\vspace{-10pt}
\subsection{Cross-Modality Masked Self-Distillation}

\begin{figure}[htbp]
    \centering \includegraphics[width=0.48\textwidth]{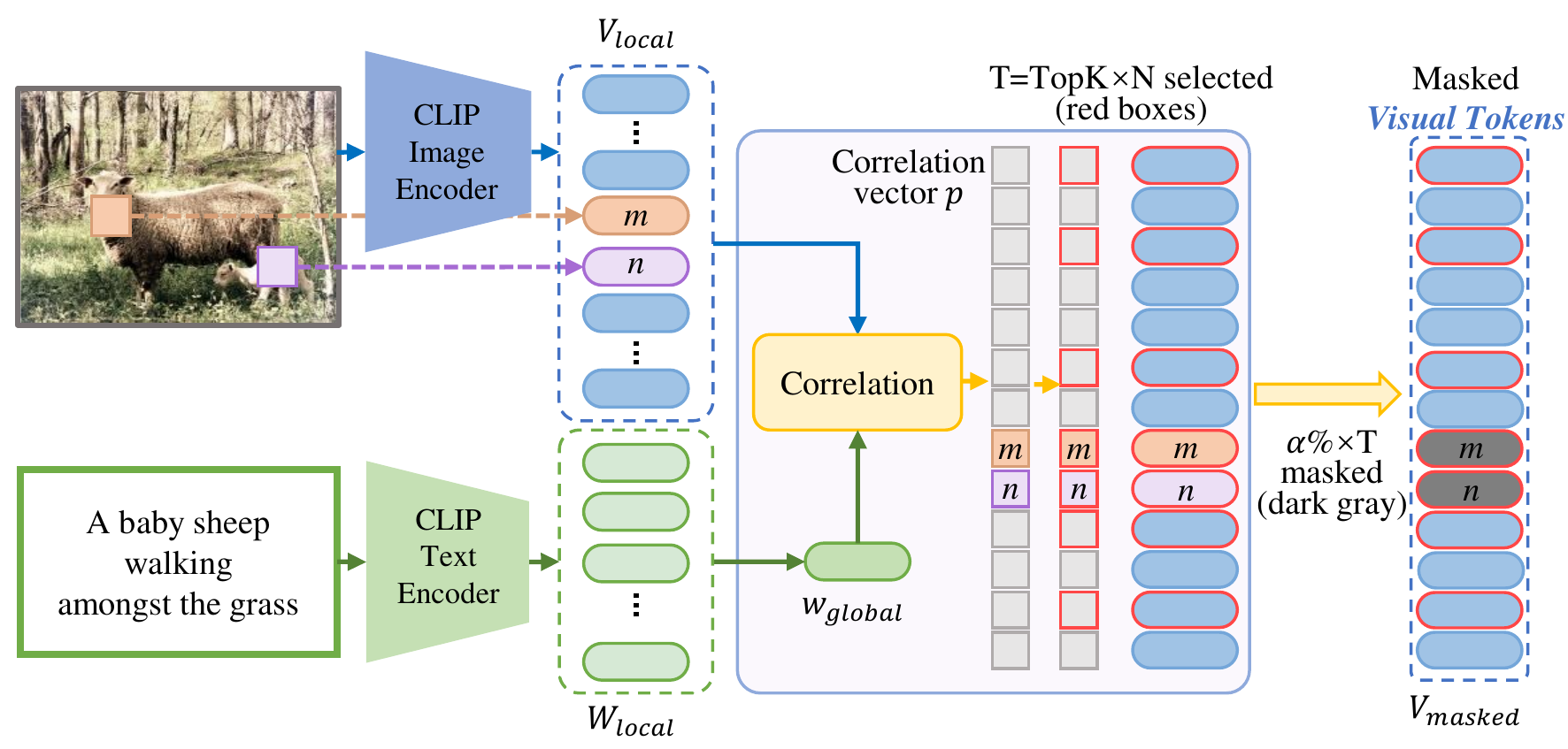}
    \vspace{-10pt}
    \caption{
    {The illustration of the introduced correlation filtering and cross-modality guided masking strategy in our language-guided masked visual self-distillation branch.}
    }
    \label{fig_correlation}
    \vspace{-15pt}
\end{figure}

Although employing the CLIP model \cite{radford2021learning} as backbone can inherit powerful image-level visual concepts from large-scale pre-training, this form of knowledge is insufficient for referring image segmentation owing to the absence of fine-grained cross-modality feature alignment. 
To better solve this issue, we propose the CM-MaskSD framework to implicitly realize dense alignment between word-level textual representations and pixel-level visual features.

\noindent \textbf{Language-Guided Masked Visual Self-Distillation (LMVSD).}
Based on the obtained visual and linguistic features in Sec. \ref{sec:mainsegmentationbranch}, we utilize the text-level representation $w_{global}$ to guide the masked visual self-distillation. 
Initially, given $w_{global} \in \mathbb{R}^{1 \times C}$ and $V_{local}=(v_{0},v_{1},\dots,v_{N-1})\in \mathbb{R}^{N \times C}$, where $v_{i}\in \mathbb{R}^{1 \times C}$, we calculate the correlation vector $p$ between them. Next, the $T$ visual feature tokens among $V_{local}$ that have the TopK highest correlation values in the correlation vector $p$ (i.e. $p^{'}$) will be selected.
{Since mis-segmentation arises due to incorrect cross-modal associations are mainly found among high-correlation visual feature tokens while visual feature tokens with inherently weak correlations can not significantly affect the final segmentation results, we anticipate that the segmentation results before and after masking off erroneously strongly-associated tokens will remain consistent in our following masked self-distillation design.
After the correlation filtering, only the visual tokens $p^{'}$ with relatively high correlations are retained and the relatively low-correlation tokens are abandoned, making the the subsequent masking strategy more effective, which can be expressed as follows:}
\begin{equation}
    \begin{split}
    p_{i} &= w_{global} \cdot v_{i}, \text{ where } i=0,1,\dots,N-1 \\
    p &= (p_{0},p_{1},...,{p^{'}_{1}},...,{p^{'}_{2}},...,{p^{'}_{T}},...,p_{N-1})
    \end{split}
    \label{eq:correlation1}
\end{equation}
where $\cdot$ denotes the dot product, $T=TopK \times N$, {$TopK$ and $T$ refer to the operation of directly selecting top $K\%$ of visual feature tokens with the highest correlation values and the number of selected visual tokens after correlation filtering}.

As shown in Fig.~\ref{fig_correlation}, these $T$ visual tokens can be categorized into two groups: (1) feature tokens that do not match the expectation (\ie the specific visual tokens are essentially not correlated with the current text), which may mislead the entire model and further induce over-segmentation (\eg visual tokens of the adult sheep, as the $m_{th}$ visual token highlighted with \textcolor{orange}{orange} color in Fig.~\ref{fig_correlation}) and (2) feature tokens that match our expectations (\ie the specific visual tokens are strongly correlated with the current text in essence) and correlate densely with the given expression (\eg visual tokens of the baby sheep, as the $n_{th}$ visual token highlighted with \textcolor{violet}{purple} color in Fig.~\ref{fig_correlation}), which directly contributes to the model's segmentation accuracy.
Then we randomly mask the selected $T$ feature tokens according to a suitable masking ratio $\alpha\%$, which involves replacing them with randomly initialized learnable tokens $v^{'}$.
\begin{equation}
    V_{masked}=(v_{0},...,{v^{'}_{1}},...,{v^{'}_{2}},...,{v^{'}_{\alpha\% \times T}},...,v_{N-1})
    \label{eq:mask1}
\end{equation}

When the first type of visual tokens are masked, we apply a self-distillation loss $Loss_{LMVSD}$ to ensure that the corresponding segmentation result $M_{LMVSD}$ is consistent with the output prediction of main segmentation branch. 
In this way, the model can associate the accurate fine-grained visual-textual features more closely and at the same time draw a clear line between the visual-textual features that should not produce strong associations. 
Therefore, the model will not be misguided by other misleading textual information, greatly avoiding the model from over-segmentation.
Simultaneously, when the second type of visual feature tokens are masked, we also use $Loss_{LMVSD}$ to ensure that the segmentation result $M_{LMVSD}$ of this branch is consistent with the output prediction of main segmentation branch. 
Since masking the corresponding tokens makes it harder for the visual distillation branch to predict segmentation result that is consistent with the main branch, this design can essentially guide the model to generate more complete segmentation masks, implicitly preventing the model from under-segmentation. 
With this language-guided masked visual self-distillation branch, the model can realize a more fine-grained alignment between the textual global features and the visual tokens than before.

To predict the segmentation results $M_{LMVSD}$, we feed the masked visual tokens $V_{masked}$ into the combination of the CLIP image encoder, neck and vision-language decoder, which stays the same architecture as the main segmentation branch. 
To be noticed, the parameters of all these three components are shared with the main segmentation branch to pursue a parameter-efficient architecture.
\begin{equation}
    \begin{split}
    V^{'}_{local}, v^{'}_{global} &= \text{CLIP Image Encoder}(V_{masked})\\
    F_{LMVSD} &= \text{Neck}(V^{'}_{local}, w_{global}) \\
    M_{LMVSD} &= \text{Vision-Language Decoder}(F_{LMVSD}, W_{local})
    \end{split}
    \label{eq:m1}
\end{equation}

\noindent \textbf{Visual-Guided Masked Language Self-Distillation (VMLSD).}
Symmetric to LMVSD elaborated above, the acquired global visual feature $v_{global}$ is utilized to guide the masked language self-distillation process in our proposed VMLSD. 
Given $v_{global} \in \mathbb{R}^{1 \times C}$ and $W_{local}=(w_{0},w_{1},\dots,w_{N-1})$, where $w_{i}\in \mathbb{R}^{1 \times C}, W_{local}\in \mathbb{R}^{N \times C}$, we calculate the correlation vector $q$ between $v_{global}$ and each $w_i$, which measures the image-word similarity. 
\begin{equation}
    \begin{split}
    q_{i} &= v_{global} \cdot w_{i}, \text{ where } i=0,1,\dots,N-1 \\
    q &= (q_{0},q_{1},...,{q^{'}_{1}},...,{q^{'}_{2}},...,{q^{'}_{T}},...,q_{N-1})
    \end{split}
    \label{eq:correlation2}
\end{equation}
where $T=TopK \times N$.
Then, the $T$ textual tokens in $W_{local}$ that have the TopK highest correlation values in $q$ (i.e. $q^{'}$) are selected by our correlation filtering operation, and $\alpha\%$ of them are randomly masked by randomly initialized learnable tokens $w^{'}$ which have the same shape as $w_{i}$. 
\begin{equation}
    W_{masked}=(w_{0},...,{w^{'}_{1}},...,{w^{'}_{2}},...,{w^{'}_{\alpha\% \times T}},...,w_{N-1})
    \label{eq:mask2}
\end{equation}

Through this process, the masked textual tokens $W_{masked}$ are obtained and fed into the combination of the CLIP text encoder, neck and vision-language decoder that stays the same architecture as main segmentation branch, to predict the segmentation results $M_{VMLSD}$. Similarly, the parameters of the CLIP text encoder, neck and vision-language decoder in VMLSD branch are all shared with the main branch to ensure the model efficiency. Benefiting from this visual-guided masked language self-distillation branch, the whole network can realize a more fine-grained alignment between the visual global features and the textual word tokens than previous.
\begin{equation}
    \begin{split}
    W^{'}_{local}, w^{'}_{global} &= \text{CLIP Image Encoder}(W_{masked})\\
    F_{VMLSD} &= \text{Neck}(W^{'}_{local}, v_{global}) \\
    M_{VMLSD} &= \text{Vision-Language Decoder}(F_{VMLSD}, V_{local})
    \end{split}    
    \label{eq:m2}
\end{equation}
\vspace{-20pt}

\subsection{Network Optimization}
During training, a compound loss function $Loss_{total}$ is adopted to constrain our model to align the word-level representations with the relevant pixel-level visual features. 
The compound loss consists of three components: (1) $Loss_{seg}$, which following CRIS \cite{wang2022cris} is a binary cross entropy (BCE) loss used to optimize the main segmentation branch and ensure accurate segmentation results; (2) $Loss_{LMVSD}$, a binary cross entropy loss, which constrains the consistency between the output of the LMVSD branch and the segmentation results output by the main segmentation branch $M_{seg}$; (3) $Loss_{VMLSD}$, similar to $Loss_{LMVSD}$, which constrains the consistency between the output of the VMLSD branch and the final segmentation results $M_{seg}$. $Loss_{LMVSD}$ and $Loss_{VMLSD}$ jointly ensure a more dense alignment between the referring expression words and image patches, implicitly preventing the whole model from over-segmentation and under-segmentation. 
The total loss is defined as follows:
\begin{equation}
    \begin{split}
    Loss_{seg} &= BCE Loss(M_{seg},GT)  \\
    Loss_{LMVSD} &= BCE Loss(M_{LMVSD},M_{seg}) \\
    Loss_{VMLSD} &= BCE Loss(M_{VMLSD},M_{seg}) \\
    Loss_{total} &= Loss_{seg}+\lambda_{1} Loss_{LMVSD}+\lambda_{2} Loss_{VMLSD}
    \end{split}
    \label{eq:loss}
\end{equation}
Here, GT denotes ground truth. $\lambda_{1}$ and $\lambda_{2}$ are hyper-parameters that control the relative importance of the two self-distillation losses. By optimizing this joint loss $Loss_{total}$, our model can learn to generate more accurate and fine-grained segmentation results that are precisely aligned with the referring expressions.
To be noticed, during the inference phase, only the main segmentation branch is reserved to obtain the corresponding segmentation masks.

\section{Experimental Results}

To evaluate the designing rationale of our method, comprehensive experiments are conducted on three benchmark datasets, including RefCOCO, RefCOCO+ and G-Ref.

\begin{table*}[htbp]
\small
    \setlength{\belowcaptionskip}{1.0pt}
    \begin{center}
    \caption{\textbf{Comparisons with the state-of-the-art approaches on three referring image segmentation benchmark datasets.}
    ``$\star$'' denotes the post-processing of DenseCRF \cite{krahenbuhl2011efficient}.
    ``-'' denotes that the result is not provided.
    The evaluation metric is mIoU.}
    \vspace{-5pt}
    \setlength{\tabcolsep}{1.6mm}{
    \begin{tabular}{l|c|c|ccc|ccc|cc}
        \toprule[1.2pt] 
        \multirow{2}{*}{Method} & \multirow{2}{*}{\makecell[c]{Vision\\Backbone}} & \multirow{2}{*}{\makecell[c]{Language\\Encoder}} & \multicolumn{3}{c|}{RefCOCO} & \multicolumn{3}{c|}{RefCOCO+} & \multicolumn{2}{c}{G-Ref} \\
        \cline{4-11}
        ~ & ~ & ~ & val & test A & test B & val & test A & test B & val & test \\
        \midrule[1.2pt]
        RMI$^\star$ \cite{liu2017recurrent}    & ResNet-101 & LSTM & 45.18 & 45.69 & 45.57 & 29.86 & 30.48 & 29.50 & - & - \\
        DMN \cite{margffoy2018dynamic}         & ResNet-101 & SRU & 49.78 & 54.83 & 45.13 & 38.88 & 44.22 & 32.29 & - & - \\
        RRN$^\star$ \cite{li2018referring}     & ResNet-101 & LSTM & 55.33 & 57.26 & 53.95 & 39.75 & 42.15 & 36.11 & - & - \\
        MAttNet \cite{yu2018mattnet}           & ResNet-101 & Bi-LSTM & 56.51 & 62.37 & 51.70 & 46.67 & 52.39 & 40.08 & 47.64 & 48.61 \\
        CMSA$^\star$ \cite{ye2019cross}        & ResNet-101 & None & 58.32 & 60.61 & 55.09 & 43.76 & 47.60 & 37.89 & - & - \\
        QRN \cite{shi2020query}           & ResNet-101 & LSTM & 59.75 & 60.96 & 58.77 & 48.23 & 52.65 & 40.89 & 42.11 & - \\
        BCAN$^\star$ \cite{hu2020bi}           & ResNet-101 & LSTM & 61.35 & 63.37 & 59.57 & 48.57 & 52.87 & 42.13 & - & - \\
        CMPC$^\star$ \cite{huang2020referring} & ResNet-101 & LSTM & 61.36 & 64.53 & 59.64 & 49.56 & 53.44 & 43.23 & - & - \\
        LSCM$^\star$ \cite{hui2020linguistic}      & ResNet-101 & LSTM & 61.47 & 64.99 & 59.55 & 49.34 & 53.12 & 43.50 & - & - \\
        MCN \cite{luo2020multi}                & DarkNet-53 & Bi-GRU & 62.44 & 64.20 & 59.71 & 50.62 & 54.99 & 44.69 & 49.22 & 49.40 \\
        CGAN \cite{luo2020cascade}             & DarkNet-53 & Bi-GRU & 64.86 & 68.04 & 62.07 & 51.03 & 55.51 & 44.06 & 51.01 & 51.69 \\
        EFNet \cite{feng2021encoder}           & ResNet-101 & Bi-GRU & 62.76 & 65.69 & 59.67 & 51.50 & 55.24 & 43.01 & - & - \\
        LTS \cite{jing2021locate}              & DarkNet-53 & Bi-GRU & 65.43 & 67.76 & 63.08 & 54.21 & 58.32 & 48.02 & 54.40 & 54.25 \\
        VLT \cite{ding2021vision}                 & DarkNet-53 & Bi-GRU & 65.65 & 68.29 & 62.73 & 55.50 & 59.20 & 49.36 & 52.99 & 56.65 \\
        {DenseCLIP \cite{rao2022denseclip}}         & {CLIP-ViT-Base}  & {CLIP} & {61.88} & {65.82} & {56.40} & {42.53} & {47.29} & {36.07} & {-} & {-} \\
        ReSTR \cite{kim2022restr}              & ViT-Base  & Transformer & 67.22 & 69.30 & 64.45 & 55.78 & 60.44 & 48.27 & - & - \\
        SeqTR \cite{zhu2022seqtr}              & DarkNet-53 & Bi-GRU & 67.26 & 69.79 & 64.12 & 54.14 & 58.93 & 48.19 & 55.67 & 55.64 \\
        CRIS \cite{wang2022cris}               & CLIP-RN101  & CLIP & 70.47 & 73.18 & 66.10 & 62.27 & 68.08 & 53.68 & 59.87 & 60.36 \\
        RefTr \cite{li2021referring}           & ResNet-101 & BERT-Base & 70.56 & 73.49 & 66.57 & 61.08 & 64.69 & 52.73 & 58.73 & 58.51 \\
        LAVT \cite{yang2022lavt}               & Swin-Base & BERT-Base & 72.73 & 75.82 & 68.79 & 62.14 & 68.38 & 55.10 & 61.24 & 62.09 \\
        \midrule
        CM-MaskSD (Ours)                            & CLIP-ViT-Base  & CLIP 
        & 72.18 & 75.21 & 67.91 
        & 64.47 & 69.29 & 56.55 & 62.67 & 62.69 \\
        CM-MaskSD (Ours)                            & CLIP-ViT-Large & CLIP & \textbf{74.89} & \textbf{77.54} & \textbf{71.28} & \textbf{67.47} & \textbf{71.80} & \textbf{59.91} & \textbf{66.53} & \textbf{66.63} \\
        \bottomrule[1.2pt]
    \end{tabular}
    \label{tab:sota}}
    \end{center}
    \vspace{-15pt}
\end{table*}

\vspace{-5pt}
\subsection{Datasets}

\textbf{RefCOCO} \cite{yu2016modeling} is one of the largest and most commonly used datasets collected from the MSCOCO \cite{lin2014microsoft} for RIS task, including 142,209 annotated expressions (average length of 3.6 words) for 50,000 objects in 19,994 images, which is split into training, validation, test A, and test B with 120,624, 10,834, 5,657 and 5,095 samples respectively.

\textbf{RefCOCO+} \cite{yu2016modeling} dataset contains 141,564 language expressions (average length of 3.5 words) with 49,856 objects in 19,992 images, which is separately split into training, validation, test A, and test B with 120,624, 10,758, 5,726, and 4,889 samples.
Compared to RefCOCO, some language expressions with absolute location descriptions are deleted in RefCOCO+ dataset, which makes it more challenging for RIS.

\textbf{G-Ref} \cite{G-Ref}, as the third benchmark dataset, includes 104,560 referring expressions (average length of 8.4 words) for 54,822 objects in 26,711 images.
It collects the language expressions from Amazon Mechanical Turk, which is different from the former two datasets. 
Following previous work, the standard UMD partition \cite{hu2016segmentation} is adopted for evaluation.

\vspace{-5pt}
\subsection{Implementation Details}

\noindent \textbf{Experimental Setup.} Our proposed framework is implemented based on Pytorch \cite{paszke2019pytorch} and trained with Tesla V100 GPUs. 
Considering the crucial scalability and the ease of implementation, the Vision Transformer \cite{dosovitskiy2020image} is adopted as the image encoder for all the experiments.
The text and image encoder are initialized by CLIP \cite{radford2021learning}, while the rest part of model weights are randomly initialized. 
During training, the input images are resized to the resolution
of 576×576 and 532×532 for ViT-Base and ViT-Large respectively
for experimental comparisons with previous SOTA methods.
To make an efficient and fair comparison, the resolution of 480×480 is adopted for ablation study.
The Adam optimizer with 32 batch size and a weight decay of 0.0005 are adopted to train the model for 100 epochs. 
With a warm-up strategy for 10 epochs during training, the initial learning rate is set to 0.00001 with a cosine decay schedule.
Following CRIS \cite{wang2022cris}, due to the extra [SOS] and [EOS] tokens, the input sentences are set with a maximum
sentence length of 17 for RefCOCO and RefCOCO+, 22 for G-Ref. 
During inference, the predicted results by our method is upsampled back to the original image size and binarized with a threshold of 0.35 to the final segmentation result. 
Any extra post-processing operations can be exploited to further boost the segmentation accuracy of our framework, but are not employed in this work.

\noindent \textbf{Evaluation Metrics.} To evaluate our proposed method, we adopt mean Intersection-over-Union (mIoU) and Precision@X as evaluation metrics. The mIoU measures the ratio between the intersection area and the union area of the prediction and ground truth among the test samples. The Precision@X measures the percentage of test samples with an IoU score higher than the threshold X that ranges from 0.5 to 0.9 with an interval of 0.1. The Precision@X focuses on the location ability of different methods.

\vspace{-5pt}
\subsection{Main Results}

\noindent \textbf{Quantitative Analysis.} To validate the superiority of our CM-MaskSD and make a fair comparison, our method is evaluated against the SOTA methods on RefCOCO \cite{yu2016modeling}, RefCOCO+ \cite{yu2016modeling} and G-Ref \cite{G-Ref} datasets. 
As presented in Table \ref{tab:sota}, our method outperforms the previous methods in terms of segmentation accuracy across all evaluation subsets of the three benchmark datasets. 
With ViT-Base \cite{dosovitskiy2020image} used as visual backbone and Transformer \cite{vaswani2017attention} as text encoder that are initialized by CLIP \cite{radford2021learning}, our method obtains comparable or better performance compared with the latest SOTA method LAVT \cite{yang2022lavt}, especially showing dominant superiority on RefCOCO+ and G-Ref datasets.
In comparison with CRIS \cite{wang2022cris} which also employs the CLIP model \cite{radford2021learning} as vision backbone and language encoder, significant performance improvement is achieved by our CM-MaskSD across all three benchmark datasets.
Besides, to further explore the potential of our framework, ViT-Large initialized by CLIP \cite{radford2021learning} is further introduced as a stronger visual backbone.
It can be clearly seen that the employment of ViT-Large as vision backbone enables our approach to further push the SOTA results.
Specifically, on RefCOCO dataset, our method achieves much higher SOTA results compared to LAVT, resulting in a significant margin of 2.16\%, 1.72\%, and 2.49\% mIoU on the validation, testA and testB sets respectively. 
On G-Ref dataset, our method surpasses the previous SOTA method LAVT \cite{yang2022lavt} by a large gap (\ie 5.29\% and 4.54\% mIoU on the validation set and test set respectively). 
The same promising results can also be found on RefCOCO+ dataset.
It is noteworthy that, instead of utilizing more powerful backbone (\eg Swin Transformer) that is pre-trained through strongly supervised manner as feature extractor like LAVT \cite{yang2022lavt}, our CM-MaskSD solely takes advantage of the simple yet effective ViT structure pre-trained through unsupervised manner \cite{radford2021learning}, not to mention that our method possesses strong scalability and doesn't contain any complex designs. 
Based on the above analysis, all the results convincingly demonstrate that our CM-MaskSD framework can better accomplish the RIS task by pursuing finer-grained cross-modality feature alignment.

\begin{figure}[t]
    \centering
    \includegraphics[width=0.45\textwidth]{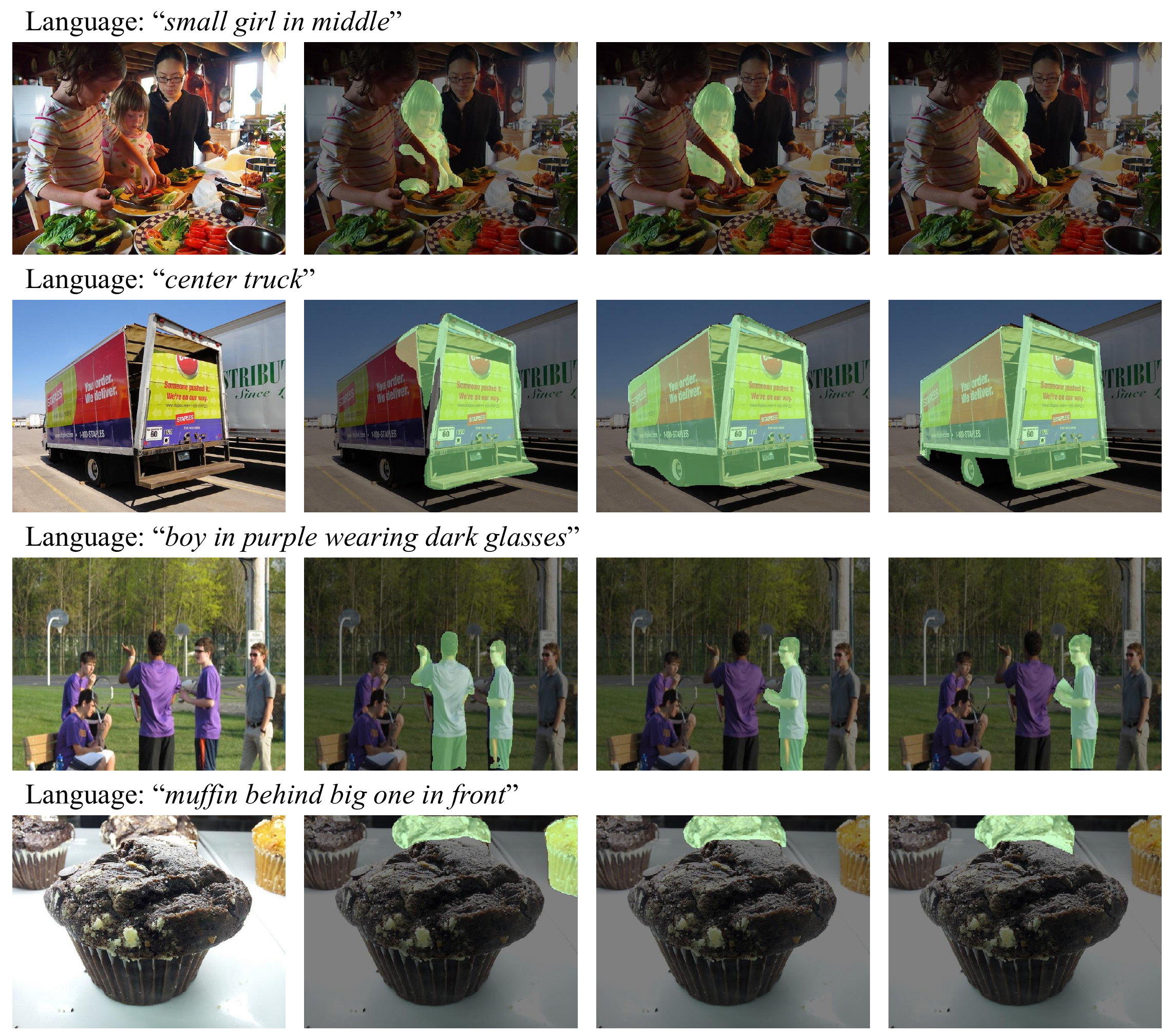}
    \begin{tabu} to 0.90\linewidth{X[1.0c] X[1.0c] X[1.0c] X[1.0c]} 
        \scriptsize{(a) Image} &  \scriptsize{(b) CRIS} &  \scriptsize{\textbf{(c) Ours}} &  \scriptsize{(d) GT} \\
    \end{tabu}
    \vspace{-5pt}
    \caption{The visual comparison of segmentation results on RefCOCO validation set. (a) input image. (b) CRIS. (c) our CM-MaskSD. (d) ground truth.}
    \label{fig_SOTAComparison_vis}
    \vspace{-15pt}
\end{figure}

\noindent \textbf{Qualitative Analysis.} In addition, 
CRIS \cite{wang2022cris} and our CM-MaskSD are further adopted for qualitative comparison since they both attempt to explore the powerful knowledge of CLIP model \cite{radford2021learning} for RIS. The visualizations in Fig.~\ref{fig_SOTAComparison_vis} convincingly illustrate that our method can accomplish RIS task more accurately and generate much better fine-grained segmentation masks of corresponding objects, greatly reducing errors caused by over-segmentation and under-segmentation.

\vspace{-5pt}
\subsection{Ablation Studies}
We conduct extensive experiments to justify the effectiveness of our design choice on RefCOCO validation set. 
ViT-Base initialized by CLIP is adopted for all the ablation study.

\begin{figure}[htbp]
    \centering
    \includegraphics[width=0.45\textwidth]{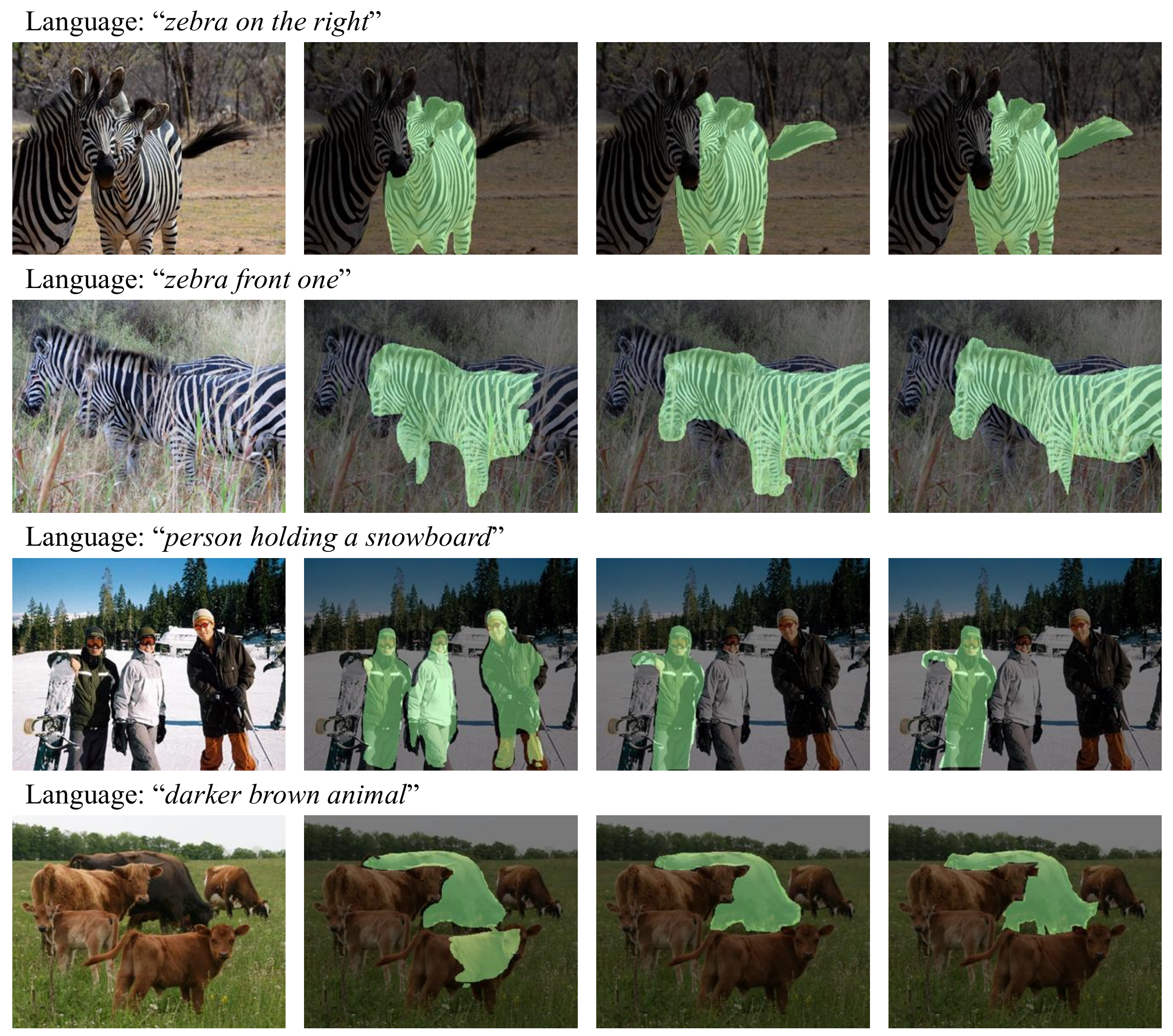}
    \begin{tabu} to 0.90\linewidth{X[1.0c] X[1.0c] X[1.0c] X[1.0c]} 
        \scriptsize{(a) Image} &  \scriptsize{(b) Baseline} &  \scriptsize{\textbf{(c) Ours}} &  \scriptsize{(d) GT} \\
    \end{tabu}
    \vspace{-5pt}
    \caption{Qualitative analysis for ablation study on our masked self-distillation design. (a) input image. (b) baseline. (c) our CM-MaskSD. (d) ground truth.}
    \label{fig_FeatureAlignment_vis}
    \vspace{-15pt}
\end{figure}

\noindent \textbf{Design of Masked Self-Distillation Manner.}
We first probe into the rationale of the proposed masked self-distillation design. 
As presented in Table \ref{tab:distillation_manners}, taking ViT-Base as the vision backbone, the baseline model obtains 70.45\% mIoU score on RefCOCO validation set. 
Either adding a single LMVSD branch or VMLSD branch to the whole architecture consistently leads to an considerable accuracy increase (1.00\% and 0.53\% mIoU). 
Since the sequence length of visual tokens is much higher than the textual sequence and the RIS task is visually oriented in essence, the experimental results that masked visual self-distillation branch can bring higher performance improvements is clearly reasonable. 
In addition, by jointly employing both LMVSD and VMLSD branches in a nearly parameter-free and strictly computation-free manner via sharing weights, our method attains 1.23\% improvements against the baseline. 
The above results fully demonstrate the benefit of exploiting the masked self-distillation framework for densely aligning linguistic and visual features. 
Besides, to show the advantage and designing rationale of our framework, we also present the visual comparison of the segmentation results in Fig. \ref{fig_FeatureAlignment_vis}.
It is clear that, compared with the baseline, our method can better solve the segmentation problem of under-segmentation and over-segmentation, benefiting from the introduced bilateral masked self-distillation branches.

\vspace{-5pt}
\begin{table}[htbp]
    \setlength{\belowcaptionskip}{1.0pt}
    \begin{center}
    \caption{Ablation study on the design of masked self-distillation manner.}
    \vspace{-5pt}
    \setlength{\tabcolsep}{0.3mm}{
    \begin{tabular}{c|c|c|c|c|c|c|c}
        \toprule[1.2pt]
        LMVSD & VMLSD & Pr@0.5 & Pr@0.7 & Pr@0.9 & mIoU & Params(M) & FLOPs(G)\\
        \midrule[1.2pt]
        - & - & 82.73 & 70.82 & 17.99 & 70.45 & 207 & 165.96\\
        \checkmark & - & 83.90 & 73.54 & 20.08 & 71.45 & 266 & 165.96\\
        - & \checkmark & 83.09 & 71.77 & 18.64 & 70.98 & 264 & 165.96\\
        \rowcolor{gray!18} \checkmark & \checkmark & 83.29 & 73.15 & 21.45 & 71.68 & 210 & 165.96 \\
        \bottomrule[1.2pt]
    \end{tabular}
    \label{tab:distillation_manners}}
    \vspace{-10pt}
    \end{center}
\end{table}

\begin{table}[htbp]
    \setlength{\belowcaptionskip}{1.0pt}
    \begin{center}
    \caption{Ablation study on language-guided masked visual self-distillation.}
    \vspace{-5pt}
    \setlength{\tabcolsep}{0.5mm}{
     \begin{tabular}{c|c|c|c|c|c|c}
          \toprule[1.2pt]
          \makecell[c]{Masking\\Ratio $\alpha$} & TopK & \makecell[c]{Loss\\Weight $\lambda_{1}$} & Pr@0.5 & Pr@0.7 & Pr@0.9 & mIoU\\
          \bottomrule[1.2pt]
          \multicolumn{7}{l}{ \textbf{(a)} Loss weight $\lambda_{1}$} \\
          \toprule[1.2pt]
          0.25 & 0.5 & 0.1 & 83.20 & 71.88 & 19.40 & 70.93 \\
          0.25 & 0.5 & 0.25 & 83.73 & 73.03 & 19.74 & 71.36 \\
          0.25 & 0.5 & 0.5 & 83.51 & 72.61 & 19.13 & 71.17 \\
          \rowcolor{gray!18} 0.25 & 0.5 & 0.75 & 83.90 & 73.54 & 20.08 & 71.45 \\
          0.25 & 0.5 & 1.0 & 82.63 & 72.53 & 20.79 & 70.72 \\
          \bottomrule[1.2pt]
          \multicolumn{7}{l}{ \textbf{(b)} Mask ratio $\alpha$} \\
          \toprule[1.2pt]
          0.1 & 0.5 & 0.75 & 83.18 & 73.39 & 19.83 & 71.12 \\
          \rowcolor{gray!18} 0.25 & 0.5 & 0.75 & 83.90 & 73.54 & 20.08 & 71.45 \\
          0.5 & 0.5 & 0.75 & 83.37 & 73.34 & 20.27 & 71.24 \\
          0.75 & 0.5 & 0.75 & 82.78 & 72.64 & 20.00 & 70.79 \\
          \bottomrule[1.2pt]
          \multicolumn{7}{l}{ \textbf{(c)} TopK } \\
          \toprule[1.2pt]
          0.25 & 0.25 & 0.75 & 83.21 & 73.41 & 20.61 & 71.09 \\
          \rowcolor{gray!18} 0.25 & 0.5 & 0.75 & 83.90 & 73.54 & 20.08 & 71.45  \\
          0.25 & 0.75 & 0.75 & 82.88 & 73.13 & 19.68 & 70.93 \\
          0.25 & 1.0 & 0.75 & 83.08 & 73.13 & 20.77 & 71.01 \\
          \bottomrule[1.2pt]
       \end{tabular}    \label{tab:Language_guided_Masked_Visual_Distillation}}
       \vspace{-20pt}
    \end{center}
\end{table}

\noindent \textbf{Language-Guided Masked Visual Self-Distillation.}
Next, we explore the potential of our proposed LMVSD branch with different hyper-parameter settings.
Table \ref{tab:Language_guided_Masked_Visual_Distillation} shows the ablation results of masking ratio $\alpha$, TopK, and loss weight $\lambda_{1}$ of LMVSD branch. 
We sequentially explore the effects of these three hyper-parameters.
Starting with loss weight $\lambda_{1}$, five diverse settings (\ie 0.1, 0.25, 0.5, 0.75, 1.0) are selected.
It is clear that $\lambda_{1}=0.75$ enables the combination of baseline and our LMVSD branch to achieve the best performance. 
On this basis, lower $\lambda_{1}$ would result in inadequate self-distillation effect and would not take full advantage of the masked self-distillation design, while higher $\lambda_{1}$ leads to an adverse effect to the overall compound loss function during optimization process.
Then we analyze the effect of different values of masking ratio $\alpha$ and TopK. 
Quantitative results in Table \ref{tab:Language_guided_Masked_Visual_Distillation} show that with masking ratio set as 0.25 and TopK set as 0.5, the introduction of LMVSD branch leads to the best model performance. 
Higher masking ratio increases the difficulty of extracting visual features and generating the same segmentation masks as the main segmentation branch during self-distillation process, while lower masking ratio could cause certain information redundancy which hinders the whole structure from unleashing the power of masked self-distillation pipeline. 
Besides, if the TopK is set too high, our introduced correlation filtering may not effectively filter out relatively irrelevant visual tokens.
On the contrary, if TopK is set too low, the range of the selected visual tokens after correlation filtering will be limited and much of the highly correlated visual tokens may be lost.
Noticeably, if our proposed correlation filtering is not performed (i.e., setting TopK=$100\%$), the model's performance consistently declines as expected, which fully proves the effectiveness of our correlation filtering insight. 

\vspace{-10pt}
\begin{table}[htbp]
    \setlength{\belowcaptionskip}{1.0pt}
    \begin{center}
    \caption{Ablation study on visual-guided masked language self-distillation.}
    \vspace{-5pt}
    \setlength{\tabcolsep}{0.5mm}{
     \begin{tabular}{c|c|c|c|c|c|c}
          \toprule[1.2pt]
          \makecell[c]{Masking \\Ratio $\alpha$} & TopK & \makecell[c]{Loss \\Weight $\lambda_{2}$} & Pr@0.5 & Pr@0.7 & Pr@0.9 & mIoU\\
          \bottomrule[1.2pt]
          \multicolumn{7}{l}{ \textbf{(a)} Loss weight $\lambda_{2}$} \\
          \toprule[1.2pt]
          0.1 & 0.5 & 0.05 & 82.76 & 71.78 & 18.73 & 70.58 \\
          \rowcolor{gray!18} 0.1 & 0.5 & 0.1 & 83.09 & 71.77 & 18.64 & 70.98 \\        
          0.1 & 0.5 & 0.25 & 83.02 & 71.40 & 18.69 & 70.41 \\
          0.1 & 0.5 & 0.5 & 82.34 & 70.86 & 18.52 & 70.02 \\
          0.1 & 0.5 & 0.75 & 81.12 & 69.92 & 18.17 & 69.47 \\
          0.1 & 0.5 & 1.0 & 81.07 & 70.25 & 17.76 & 69.28 \\
          \bottomrule[1.2pt]
          \multicolumn{7}{l}{ \textbf{(b)} Mask ratio $\alpha$} \\
          \toprule[1.2pt]
          0.05 & 0.5 & 0.1 & 83.26 & 71.93 & 18.52 & 70.71 \\
          \rowcolor{gray!18} 0.1 & 0.5 & 0.1 & 83.09 & 71.77 & 18.64 & 70.98 \\
          0.25 & 0.5 & 0.1 & 82.72 & 71.11 & 18.17 & 70.43 \\
          0.5 & 0.5 & 0.1 & 82.82 & 71.92 & 17.97 & 70.51 \\
          0.75 & 0.5 & 0.1 & 82.67 & 71.31 & 18.24 & 70.54 \\
          \bottomrule[1.2pt]
          \multicolumn{7}{l}{ \textbf{(c)} TopK } \\
          \toprule[1.2pt]
          0.1 & 0.25 & 0.1 & 83.22 & 72.12 & 18.18 & 70.68 \\
          \rowcolor{gray!18} 0.1 & 0.5 & 0.1 & 83.09 & 71.77 & 18.64 & 70.98 \\
          0.1 & 0.75 & 0.1 & 82.66 & 71.09 & 18.07 & 70.23 \\
          0.1 & 1.0 & 0.1 & 82.92 & 71.66 & 18.32 & 70.57 \\
          \bottomrule[1.2pt]
       \end{tabular}
    \label{tab:Visual_guided_Masked_Language_Distillation}}
    \vspace{-10pt}
    \end{center}
\end{table}

\noindent \textbf{Visual-Guided Masked Language Self-Distillation.}
To further explore the potential of our proposed VMLSD branch, similar ablation study like LMVSD branch are conducted. 
Table \ref{tab:Visual_guided_Masked_Language_Distillation} presents the experimental results.
Different from the ablation study on LMVSD branch, the baseline with our VMLSD branch inserted obtains the best model performance when loss weight $\lambda_{2}$, masking ratio $\alpha$ and TopK are set as 0.1, 0.1 and 0.5, respectively.
Since the referring image segmentation is essentially a vision-dominated task, dislike the optimal setting $\lambda_{1}=0.75$ in the LMVSD branch, the VMLSD branch with $\lambda_{2}=0.1$ yields the best segmentation accuracy.
Besides, since the visual-guided masking operation is performed at the word level, based on the textual low redundancy due to the short expression length (average 3.6 words on RefCOCO dataset), a high $\alpha$ will lead to the missing of important textual information.
Additionally, similar to LMVSD branch, either the TopK is set unsuitably high or low, the segmentation accuracy would decrease, since the employed correlation filtering can not effectively filter out irrelevant textual information or much of the strongly correlated word tokens are accordingly lost for masked self-distillation.
Furthermore, if TopK is set as $100\%$ (\ie our correlation filtering is removed), it would result in masking many unimportant regions that have minimal impact on the final segmentation results. Masking these regions and then constraining the consistency of corresponding segmentation results before and after masking (i.e. pre- and post- masking) would essentially become meaningless. Consequently, the masking operation would be highly inefficient, leading to insufficient learning of cross-modality feature alignment and thus deteriorated model performance.

\vspace{-10pt}
\begin{table}[htbp]
    \setlength{\belowcaptionskip}{1.0pt}
    \begin{center}
    \caption{Ablation study on the proposed cross-modality guided masking strategy in our masked self-distillation structure.}
    \vspace{-5pt}
    \setlength{\tabcolsep}{0.25mm}{
    \begin{tabular}{c|c|c|c|c|c|c}
    \toprule[1.2pt]
    LMVSD & VMLSD & \makecell[c]{Masking\\Operation} & Pr@0.5 & Pr@0.7 & Pr@0.9 & mIoU \\
    \midrule[1.2pt]
    - & - & - & 82.73 & 70.82 & 17.99 & 70.45   \\
    \hline
    \rowcolor{gray!18} \checkmark & - & \checkmark & 83.90 & 73.54 & 20.08 & 71.45 \\
    \checkmark & - & - & 83.06 & 71.77 & 19.00 & 70.78 \\
    \hline
    \rowcolor{gray!18} - & \checkmark & \checkmark & 83.09 & 71.77 & 18.64 & 70.98 \\
    - & \checkmark & - & 82.96 & 71.83 & 18.89 & 70.88 \\
    \hline
    \rowcolor{gray!18} \checkmark & \checkmark & \checkmark & 83.29 & 73.15 & 21.45 & 71.68 \\
    \checkmark & \checkmark & - & 82.99 & 72.28 & 19.22 & 71.12 \\
    \bottomrule[1.2pt]
  \end{tabular}
    \label{tab:Masking}}
    \end{center}
    \vspace{-10pt}
\end{table}

\noindent \textbf{Necessity of Masking Operation in Masked Self-Distillation Structure.}
Although we have explored the influence of different mask ratios on our masked self-distillation architecture, the model performance when masking operation is removed has not been fully discussed.
In this case, to prove the necessity of our proposed masking strategy, we further investigate the model performance when masking operation is not performed in the whole structure. 
As shown in Table \ref{tab:Masking}, for introducing either a single masked self-distillation branch or both of the LMVSD and VMLSD branches, removing the masking operation consistently results in a considerable decline in model accuracy, confirming the effectiveness of the proposed cross-modality guided masking strategy to better guide the whole architecture to achieve dense multimodal feature alignment.

\vspace{-10pt}
\begin{table}[htbp]
    \setlength{\belowcaptionskip}{1.0pt}
    \begin{center}
    \caption{Ablation study on different supervision schemes for parameter-efficient masked self-distillation structure.}
    \vspace{-5pt}
    \setlength{\tabcolsep}{0.3mm}{
    \begin{tabular}{c|c|c|c|c|c|c}
    \toprule[1.2pt]
    LMVSD & VMLSD & \makecell[c]{Supervision\\Manners} & Pr@0.5 & Pr@0.7 & Pr@0.9 & mIoU \\
    \midrule[1.2pt]
    - & - & - & 82.73 & 70.82 & 17.99 & 70.45   \\
    \hline
    \rowcolor{gray!18} \checkmark & - & Distillation & 83.90 & 73.54 & 20.08 & 71.45 \\
    \checkmark & - & Ground Truth & 82.89 & 71.72 & 17.76 & 70.73 \\
    \hline
    \rowcolor{gray!18} - & \checkmark & Distillation & 83.09 & 71.77 & 18.64 & 70.98 \\
    - & \checkmark & Ground Truth & 82.91 & 71.05 & 17.81 & 70.55 \\
    \hline
    \rowcolor{gray!18} \checkmark & \checkmark & Distillation & 83.29 & 73.15 & 21.45 & 71.68 \\
    \checkmark & \checkmark & Ground Truth & 83.25 & 72.72 & 18.83 & 70.95 \\
    \bottomrule[1.2pt]
  \end{tabular}
    \label{tab:Supervision_Schemes}}
    \vspace{-10pt}
    \end{center}
\end{table}

\noindent \textbf{Supervision Scheme in Masked Self-Distillation Structure.}
Additionally, we also investigate two different supervision approaches and their impact on our entire framework, including directly supervising the two distillation branches with ground truth or employing our proposed self-distillation manner. 
The quantitative results presented in Table \ref{tab:Supervision_Schemes} validate the effectiveness and rationality of our current design choice.
From the perspective of design logic, it would be natural to think of directly supervising the two distillation branches with ground truth. 
Although it is true that this supervision manner is not logically infeasible, we found that this manner is sub-optimal for our overall framework and aligns more with auxiliary loss supervision, rather than the distillation motivation underlying our cross-modality masked self-distillation framework. 
Typically, using ground-truth for supervision can more directly and effectively assist the model. 
However, in our case, where we aim to better guide cross-modality feature alignment through our proposed cross-modality guided masking operation that masks out the visually or textually misleading regions and enforces consistency between pre- and post-masking segmentation results, using ground-truth as direct supervision signal for both the two distillation branches and main segmentation branch simultaneously would make ground-truth act as an ineffective intermediate quantity. 
In contrast, our original intention is to directly enforce consistency between pre- and post-masking segmentation results, and it is more direct and effective to use the main segmentation branch's output as labels for the two self-distillation branches rather than guiding them independently towards ground truth.

\vspace{-10pt}
\begin{table}[H]
    \setlength{\belowcaptionskip}{1.0pt}
    \begin{center}
    \caption{Ablation study on sharing weights for parameter-efficient masked self-distillation structure.}
    \vspace{-10pt}
    \setlength{\tabcolsep}{0.15mm}{
    \begin{tabular}{c|c|c|c|c|c|c|c|c}
    \toprule[1.2pt]
    LMVSD & VMLSD & \makecell[c]{Sharing\\weights} & Pr@0.5 & Pr@0.7 & Pr@0.9 & mIoU & Params(M) & FLOPs(G)\\
    \midrule[1.2pt]
    - & - & - & 82.73 & 70.82 & 17.99 & 70.45 & 207 & 165.96  \\
    \hline
    \checkmark & - & - & 83.90 & 73.54 & 20.08 & 71.45 & 266 & 165.96 \\
    \checkmark & - & \checkmark & 83.47 & 73.02 & 18.53 & 71.13 & 210 & 165.96\\
    \hline
    - & \checkmark & - & 83.09 & 71.77 & 18.64 & 70.98 & 264 & 165.96 \\
    - & \checkmark & \checkmark & 83.29 & 71.80 & 18.28 & 70.88 & 207 & 165.96\\
    \hline
    \checkmark & \checkmark & - & 83.26 & 72.24 & 19.03 & 71.26 & 323 & 165.96\\
    \rowcolor{gray!18} \checkmark & \checkmark & \checkmark & 83.29 & 73.15 & 21.45 & 71.68 & 210 & 165.96\\
    \bottomrule[1.2pt]
  \end{tabular}
    \label{tab:Shared_Weights}}
    \vspace{-15pt}
    \end{center}
\end{table}

\noindent \textbf{Sharing Weights for Parameter-Efficient Distillation Structure.}
Finally, we investigate the potential of our CM-MaskSD in the case of sharing weights for a parameter-efficient structure. 
The experimental results are presented in Table \ref{tab:Shared_Weights}. 
Through sharing parameters of neck module and vision-language decoder between the main branch and self-distillation branch, the introduction of either LMVSD branch or VMLSD branch leads to the performance improvement of 0.68\% and 0.43\% mIoU respectively compared with the baseline. 
Although the resulted improvement is a little inferior to that without sharing parameters, employing the LMVSD branch or VMLSD branch in this parameter-efficient manner can consistently boost model performance with almost no additional parameters introduced.
When both of the LMVSD branch and the VMLSD branch are simultaneously introduced, it leads to a 0.81\% accuracy increase over the baseline under the circumstance of not sharing weights, which is not promising as the mIoU score 71.45\% made by solely introducing our LMVSD.
We believe it is because that not sharing weights can not fully unleash the power of our cross-modality masked self-distillation structure due to the optimization difficulties, since all the parameters of these three branches (\ie VMLSD branch, LMVSD branch and main segmentation branch) need to be optimized at the same time and there may be optimization collision between different parts. 
Thus, in order to effectively optimize the whole structure and to pursue a parameter-efficient framework, our CM-MaskSD adopts the designing scheme of sharing parameters and achieves the highest segmentation accuracy with only introduced negligible extra parameters (\ie 3M) and none extra computational costs.

\section{Conclusion and Future Work}
\label{Conclusion}

In this paper, we present the first study to explore the potential of masked multimodal modeling with self-distillation for RIS task and propose a novel framework CM-MaskSD that exploits masked multimodal modeling with self-distillation. 
It inherits transferred knowledge of image-text semantic alignment from CLIP and achieves dense feature alignment for improved segmentation accuracy. 
Our CM-MaskSD is scalable and flexible, 
among which our masked self-distillation design is essentially plug-and-play and easy-to-implement.
Extensive experiments on three RIS benchmark datasets demonstrate that our CM-MaskSD greatly outperforms previous SOTA methods with negligible introduced parameters.

Our approach provides a novel solution to better guide the model to realize fine-grained feature alignment in RIS, inspiring new research in this direction. One potential limitation could be that our framework mainly concentrates on ViT-based structures, but recent research suggests that our correlation filtering and cross-modality guided masking strategy can be accordingly adjusted to overcome this limitation. This provides a future research direction to develop a more general and powerful masked self-distillation framework that can assist various types of Transformer-based (\ie hierarchical Swin Transformer) and CNN-based models in achieving dense multimodal feature alignment.


\bibliographystyle{IEEEtran}
\bibliography{Reference}

\begin{thebibliography}{10}
\providecommand{\url}[1]{#1}
\csname url@samestyle\endcsname
\providecommand{\newblock}{\relax}
\providecommand{\bibinfo}[2]{#2}
\providecommand{\BIBentrySTDinterwordspacing}{\spaceskip=0pt\relax}
\providecommand{\BIBentryALTinterwordstretchfactor}{4}
\providecommand{\BIBentryALTinterwordspacing}{\spaceskip=\fontdimen2\font plus
\BIBentryALTinterwordstretchfactor\fontdimen3\font minus \fontdimen4\font\relax}
\providecommand{\BIBforeignlanguage}[2]{{%
\expandafter\ifx\csname l@#1\endcsname\relax
\typeout{** WARNING: IEEEtran.bst: No hyphenation pattern has been}%
\typeout{** loaded for the language `#1'. Using the pattern for}%
\typeout{** the default language instead.}%
\else
\language=\csname l@#1\endcsname
\fi
#2}}
\providecommand{\BIBdecl}{\relax}
\BIBdecl

\bibitem{hu2016segmentation}
R.~Hu, M.~Rohrbach, and T.~Darrell, ``Segmentation from natural language expressions,'' in \emph{Computer Vision--ECCV 2016: 14th European Conference, Amsterdam, The Netherlands, October 11--14, 2016, Proceedings, Part I 14}.\hskip 1em plus 0.5em minus 0.4em\relax Springer, 2016, pp. 108--124.

\bibitem{wang2022cris}
Z.~Wang, Y.~Lu, Q.~Li, X.~Tao, Y.~Guo, M.~Gong, and T.~Liu, ``Cris: Clip-driven referring image segmentation,'' in \emph{Proceedings of the IEEE/CVF conference on computer vision and pattern recognition}, 2022, pp. 11\,686--11\,695.

\bibitem{yang2022lavt}
Z.~Yang, J.~Wang, Y.~Tang, K.~Chen, H.~Zhao, and P.~H. Torr, ``Lavt: Language-aware vision transformer for referring image segmentation,'' in \emph{Proceedings of the IEEE/CVF Conference on Computer Vision and Pattern Recognition}, 2022, pp. 18\,155--18\,165.

\bibitem{kim2022restr}
N.~Kim, D.~Kim, C.~Lan, W.~Zeng, and S.~Kwak, ``Restr: Convolution-free referring image segmentation using transformers,'' in \emph{Proceedings of the IEEE/CVF Conference on Computer Vision and Pattern Recognition}, 2022, pp. 18\,145--18\,154.

\bibitem{hochreiter1997long}
S.~Hochreiter and J.~Schmidhuber, ``Long short-term memory,'' \emph{Neural computation}, vol.~9, no.~8, pp. 1735--1780, 1997.

\bibitem{shi2018key}
H.~Shi, H.~Li, F.~Meng, and Q.~Wu, ``Key-word-aware network for referring expression image segmentation,'' in \emph{Proceedings of the European Conference on Computer Vision (ECCV)}, 2018, pp. 38--54.

\bibitem{margffoy2018dynamic}
E.~Margffoy-Tuay, J.~C. P{\'e}rez, E.~Botero, and P.~Arbel{\'a}ez, ``Dynamic multimodal instance segmentation guided by natural language queries,'' in \emph{Proceedings of the European Conference on Computer Vision (ECCV)}, 2018, pp. 630--645.

\bibitem{li2018referring}
R.~Li, K.~Li, Y.-C. Kuo, M.~Shu, X.~Qi, X.~Shen, and J.~Jia, ``Referring image segmentation via recurrent refinement networks,'' in \emph{Proceedings of the IEEE Conference on Computer Vision and Pattern Recognition}, 2018, pp. 5745--5753.

\bibitem{qiu2019referring}
S.~Qiu, Y.~Zhao, J.~Jiao, Y.~Wei, and S.~Wei, ``Referring image segmentation by generative adversarial learning,'' \emph{IEEE Transactions on Multimedia}, vol.~22, no.~5, pp. 1333--1344, 2019.

\bibitem{ye2020dual}
L.~Ye, Z.~Liu, and Y.~Wang, ``Dual convolutional lstm network for referring image segmentation,'' \emph{IEEE Transactions on Multimedia}, vol.~22, no.~12, pp. 3224--3235, 2020.

\bibitem{shi2020query}
H.~Shi, H.~Li, Q.~Wu, and K.~N. Ngan, ``Query reconstruction network for referring expression image segmentation,'' \emph{IEEE Transactions on Multimedia}, vol.~23, pp. 995--1007, 2020.

\bibitem{lin2021structured}
L.~Lin, P.~Yan, X.~Xu, S.~Yang, K.~Zeng, and G.~Li, ``Structured attention network for referring image segmentation,'' \emph{IEEE Transactions on Multimedia}, vol.~24, pp. 1922--1932, 2021.

\bibitem{liu2017recurrent}
C.~Liu, Z.~Lin, X.~Shen, J.~Yang, X.~Lu, and A.~Yuille, ``Recurrent multimodal interaction for referring image segmentation,'' in \emph{Proceedings of the IEEE international conference on computer vision}, 2017, pp. 1271--1280.

\bibitem{yu2018mattnet}
L.~Yu, Z.~Lin, X.~Shen, J.~Yang, X.~Lu, M.~Bansal, and T.~L. Berg, ``Mattnet: Modular attention network for referring expression comprehension,'' in \emph{Proceedings of the IEEE conference on computer vision and pattern recognition}, 2018, pp. 1307--1315.

\bibitem{huang2020referring}
S.~Huang, T.~Hui, S.~Liu, G.~Li, Y.~Wei, J.~Han, L.~Liu, and B.~Li, ``Referring image segmentation via cross-modal progressive comprehension,'' in \emph{Proceedings of the IEEE/CVF conference on computer vision and pattern recognition}, 2020, pp. 10\,488--10\,497.

\bibitem{ye2019cross}
L.~Ye, M.~Rochan, Z.~Liu, and Y.~Wang, ``Cross-modal self-attention network for referring image segmentation,'' in \emph{Proceedings of the IEEE/CVF conference on computer vision and pattern recognition}, 2019, pp. 10\,502--10\,511.

\bibitem{ding2021vision}
H.~Ding, C.~Liu, S.~Wang, and X.~Jiang, ``Vision-language transformer and query generation for referring segmentation,'' in \emph{Proceedings of the IEEE/CVF International Conference on Computer Vision}, 2021, pp. 16\,321--16\,330.

\bibitem{wu2022towards}
J.~Wu, X.~Li, X.~Li, H.~Ding, Y.~Tong, and D.~Tao, ``Towards robust referring image segmentation,'' \emph{arXiv preprint arXiv:2209.09554}, 2022.

\bibitem{devlin2018bert}
J.~Devlin, M.-W. Chang, K.~Lee, and K.~Toutanova, ``Bert: Pre-training of deep bidirectional transformers for language understanding,'' \emph{arXiv preprint arXiv:1810.04805}, 2018.

\bibitem{liu2019roberta}
Y.~Liu, M.~Ott, N.~Goyal, J.~Du, M.~Joshi, D.~Chen, O.~Levy, M.~Lewis, L.~Zettlemoyer, and V.~Stoyanov, ``Roberta: A robustly optimized bert pretraining approach,'' \emph{arXiv preprint arXiv:1907.11692}, 2019.

\bibitem{clark2020electra}
K.~Clark, M.-T. Luong, Q.~V. Le, and C.~D. Manning, ``Electra: Pre-training text encoders as discriminators rather than generators,'' \emph{arXiv preprint arXiv:2003.10555}, 2020.

\bibitem{dosovitskiy2020image}
A.~Dosovitskiy, L.~Beyer, A.~Kolesnikov, D.~Weissenborn, X.~Zhai, T.~Unterthiner, M.~Dehghani, M.~Minderer, G.~Heigold, S.~Gelly \emph{et~al.}, ``An image is worth 16x16 words: Transformers for image recognition at scale,'' \emph{arXiv preprint arXiv:2010.11929}, 2020.

\bibitem{bao2021beit}
H.~Bao, L.~Dong, S.~Piao, and F.~Wei, ``Beit: Bert pre-training of image transformers,'' \emph{arXiv preprint arXiv:2106.08254}, 2021.

\bibitem{peng2022beit}
Z.~Peng, L.~Dong, H.~Bao, Q.~Ye, and F.~Wei, ``Beit v2: Masked image modeling with vector-quantized visual tokenizers,'' \emph{arXiv preprint arXiv:2208.06366}, 2022.

\bibitem{xie2022simmim}
Z.~Xie, Z.~Zhang, Y.~Cao, Y.~Lin, J.~Bao, Z.~Yao, Q.~Dai, and H.~Hu, ``Simmim: A simple framework for masked image modeling,'' in \emph{Proceedings of the IEEE/CVF Conference on Computer Vision and Pattern Recognition}, 2022, pp. 9653--9663.

\bibitem{he2022masked}
K.~He, X.~Chen, S.~Xie, Y.~Li, P.~Doll{\'a}r, and R.~Girshick, ``Masked autoencoders are scalable vision learners,'' in \emph{Proceedings of the IEEE/CVF Conference on Computer Vision and Pattern Recognition}, 2022, pp. 16\,000--16\,009.

\bibitem{huang2022green}
L.~Huang, S.~You, M.~Zheng, F.~Wang, C.~Qian, and T.~Yamasaki, ``Green hierarchical vision transformer for masked image modeling,'' \emph{Advances in Neural Information Processing Systems}, vol.~35, pp. 19\,997--20\,010, 2022.

\bibitem{liu2022mixmim}
J.~Liu, X.~Huang, Y.~Liu, and H.~Li, ``Mixmim: Mixed and masked image modeling for efficient visual representation learning,'' \emph{arXiv preprint arXiv:2205.13137}, 2022.

\bibitem{tian2023designing}
K.~Tian, Y.~Jiang, Q.~Diao, C.~Lin, L.~Wang, and Z.~Yuan, ``Designing bert for convolutional networks: Sparse and hierarchical masked modeling,'' \emph{arXiv preprint arXiv:2301.03580}, 2023.

\bibitem{wang2023positive}
J.~Wang, S.~Qian, J.~Hu, and R.~Hong, ``Positive unlabeled fake news detection via multi-modal masked transformer network,'' \emph{IEEE Transactions on Multimedia}, 2023.

\bibitem{chen2021feature}
W.~Chen, Y.~Liu, N.~Pu, W.~Wang, L.~Liu, and M.~S. Lew, ``Feature estimations based correlation distillation for incremental image retrieval,'' \emph{IEEE Transactions on Multimedia}, vol.~24, pp. 1844--1856, 2021.

\bibitem{deng2021extended}
C.~Deng, M.~Wang, L.~Liu, Y.~Liu, and Y.~Jiang, ``Extended feature pyramid network for small object detection,'' \emph{IEEE Transactions on Multimedia}, vol.~24, pp. 1968--1979, 2021.

\bibitem{chen2021temporal}
C.~Chen, S.~Dong, Y.~Tian, K.~Cao, L.~Liu, and Y.~Guo, ``Temporal self-ensembling teacher for semi-supervised object detection,'' \emph{IEEE Transactions on Multimedia}, vol.~24, pp. 3679--3692, 2021.

\bibitem{hao2022cdfkd}
Z.~Hao, Y.~Luo, Z.~Wang, H.~Hu, and J.~An, ``Cdfkd-mfs: Collaborative data-free knowledge distillation via multi-level feature sharing,'' \emph{IEEE Transactions on Multimedia}, vol.~24, pp. 4262--4274, 2022.

\bibitem{hu2022mmnet}
L.~Hu, Z.~Zhao, X.~Ge, X.~Song, and L.~Nie, ``Mmnet: Multi-modal fusion with mutual learning network for fake news detection,'' \emph{arXiv preprint arXiv:2212.05699}, 2022.

\bibitem{yang2022masked}
Z.~Yang, Z.~Li, M.~Shao, D.~Shi, Z.~Yuan, and C.~Yuan, ``Masked generative distillation,'' in \emph{Computer Vision--ECCV 2022: 17th European Conference, Tel Aviv, Israel, October 23--27, 2022, Proceedings, Part XI}.\hskip 1em plus 0.5em minus 0.4em\relax Springer, 2022, pp. 53--69.

\bibitem{bai2023masked}
Y.~Bai, Z.~Wang, J.~Xiao, C.~Wei, H.~Wang, A.~L. Yuille, Y.~Zhou, and C.~Xie, ``Masked autoencoders enable efficient knowledge distillers,'' in \emph{Proceedings of the IEEE/CVF Conference on Computer Vision and Pattern Recognition}, 2023, pp. 24\,256--24\,265.

\bibitem{huang2022masked}
T.~Huang, Y.~Zhang, S.~You, F.~Wang, C.~Qian, J.~Cao, and C.~Xu, ``Masked distillation with receptive tokens,'' \emph{arXiv preprint arXiv:2205.14589}, 2022.

\bibitem{son2023maskedkd}
S.~Son, N.~Lee, and J.~Lee, ``Maskedkd: Efficient distillation of vision transformers with masked images,'' \emph{arXiv preprint arXiv:2302.10494}, 2023.

\bibitem{radford2021learning}
A.~Radford, J.~W. Kim, C.~Hallacy, A.~Ramesh, G.~Goh, S.~Agarwal, G.~Sastry, A.~Askell, P.~Mishkin, J.~Clark \emph{et~al.}, ``Learning transferable visual models from natural language supervision,'' in \emph{International conference on machine learning}.\hskip 1em plus 0.5em minus 0.4em\relax PMLR, 2021, pp. 8748--8763.

\bibitem{krahenbuhl2011efficient}
P.~Kr{\"a}henb{\"u}hl and V.~Koltun, ``Efficient inference in fully connected crfs with gaussian edge potentials,'' \emph{Advances in neural information processing systems}, vol.~24, 2011.

\bibitem{hu2020bi}
Z.~Hu, G.~Feng, J.~Sun, L.~Zhang, and H.~Lu, ``Bi-directional relationship inferring network for referring image segmentation,'' in \emph{Proceedings of the IEEE/CVF conference on computer vision and pattern recognition}, 2020, pp. 4424--4433.

\bibitem{hui2020linguistic}
T.~Hui, S.~Liu, S.~Huang, G.~Li, S.~Yu, F.~Zhang, and J.~Han, ``Linguistic structure guided context modeling for referring image segmentation,'' in \emph{Computer Vision--ECCV 2020: 16th European Conference, Glasgow, UK, August 23--28, 2020, Proceedings, Part X 16}.\hskip 1em plus 0.5em minus 0.4em\relax Springer, 2020, pp. 59--75.

\bibitem{luo2020multi}
G.~Luo, Y.~Zhou, X.~Sun, L.~Cao, C.~Wu, C.~Deng, and R.~Ji, ``Multi-task collaborative network for joint referring expression comprehension and segmentation,'' in \emph{Proceedings of the IEEE/CVF Conference on computer vision and pattern recognition}, 2020, pp. 10\,034--10\,043.

\bibitem{luo2020cascade}
G.~Luo, Y.~Zhou, R.~Ji, X.~Sun, J.~Su, C.-W. Lin, and Q.~Tian, ``Cascade grouped attention network for referring expression segmentation,'' in \emph{Proceedings of the 28th ACM International Conference on Multimedia}, 2020, pp. 1274--1282.

\bibitem{feng2021encoder}
G.~Feng, Z.~Hu, L.~Zhang, and H.~Lu, ``Encoder fusion network with co-attention embedding for referring image segmentation,'' in \emph{Proceedings of the IEEE/CVF Conference on Computer Vision and Pattern Recognition}, 2021, pp. 15\,506--15\,515.

\bibitem{jing2021locate}
Y.~Jing, T.~Kong, W.~Wang, L.~Wang, L.~Li, and T.~Tan, ``Locate then segment: A strong pipeline for referring image segmentation,'' in \emph{Proceedings of the IEEE/CVF Conference on Computer Vision and Pattern Recognition}, 2021, pp. 9858--9867.

\bibitem{rao2022denseclip}
Y.~Rao, W.~Zhao, G.~Chen, Y.~Tang, Z.~Zhu, G.~Huang, J.~Zhou, and J.~Lu, ``Denseclip: Language-guided dense prediction with context-aware prompting,'' in \emph{Proceedings of the IEEE/CVF Conference on Computer Vision and Pattern Recognition}, 2022, pp. 18\,082--18\,091.

\bibitem{zhu2022seqtr}
C.~Zhu, Y.~Zhou, Y.~Shen, G.~Luo, X.~Pan, M.~Lin, C.~Chen, L.~Cao, X.~Sun, and R.~Ji, ``Seqtr: A simple yet universal network for visual grounding,'' in \emph{Computer Vision--ECCV 2022: 17th European Conference, Tel Aviv, Israel, October 23--27, 2022, Proceedings, Part XXXV}.\hskip 1em plus 0.5em minus 0.4em\relax Springer, 2022, pp. 598--615.

\bibitem{li2021referring}
M.~Li and L.~Sigal, ``Referring transformer: A one-step approach to multi-task visual grounding,'' \emph{Advances in neural information processing systems}, vol.~34, pp. 19\,652--19\,664, 2021.

\bibitem{yu2016modeling}
L.~Yu, P.~Poirson, S.~Yang, A.~C. Berg, and T.~L. Berg, ``Modeling context in referring expressions,'' in \emph{Computer Vision--ECCV 2016: 14th European Conference, Amsterdam, The Netherlands, October 11-14, 2016, Proceedings, Part II 14}.\hskip 1em plus 0.5em minus 0.4em\relax Springer, 2016, pp. 69--85.

\bibitem{lin2014microsoft}
T.-Y. Lin, M.~Maire, S.~Belongie, J.~Hays, P.~Perona, D.~Ramanan, P.~Doll{\'a}r, and C.~L. Zitnick, ``Microsoft coco: Common objects in context,'' in \emph{Computer Vision--ECCV 2014: 13th European Conference, Zurich, Switzerland, September 6-12, 2014, Proceedings, Part V 13}.\hskip 1em plus 0.5em minus 0.4em\relax Springer, 2014, pp. 740--755.

\bibitem{G-Ref}
V.~K. Nagaraja, V.~I. Morariu, and L.~S. Davis, ``Modeling context between objects for referring expression understanding,'' in \emph{Computer Vision--ECCV 2016: 14th European Conference, Amsterdam, The Netherlands, October 11--14, 2016, Proceedings, Part IV 14}.\hskip 1em plus 0.5em minus 0.4em\relax Springer, 2016, pp. 792--807.

\bibitem{paszke2019pytorch}
A.~Paszke, S.~Gross, F.~Massa, A.~Lerer, J.~Bradbury, G.~Chanan, T.~Killeen, Z.~Lin, N.~Gimelshein, L.~Antiga \emph{et~al.}, ``Pytorch: An imperative style, high-performance deep learning library,'' \emph{Advances in neural information processing systems}, vol.~32, 2019.

\bibitem{vaswani2017attention}
A.~Vaswani, N.~Shazeer, N.~Parmar, J.~Uszkoreit, L.~Jones, A.~N. Gomez, {\L}.~Kaiser, and I.~Polosukhin, ``Attention is all you need,'' \emph{Advances in neural information processing systems}, vol.~30, 2017.

\end{thebibliography}

\newpage

 





\end{document}